\documentclass[letterpaper, 10 pt, conference]{ieeeconf}  

\IEEEoverridecommandlockouts                              
\usepackage{subcaption}
\usepackage{mwe}
\usepackage{graphics} 
\usepackage{bm}
\usepackage{siunitx}
\usepackage{relsize}
\usepackage{amsmath} 
\usepackage{amssymb}  
\usepackage[ruled,vlined]{algorithm2e} 
\usepackage{booktabs}
\usepackage{siunitx}
\usepackage{multirow}
\usepackage{makecell}
\usepackage{subcaption} 
\usepackage{hyperref}
\usepackage[usenames, dvipsnames,table,xcdraw]{xcolor}
\usepackage{pifont}
\UseRawInputEncoding

\usepackage[capitalise]{cleveref}
\usepackage{bm}
\usepackage{mathtools}

\newcommand{\rsec}                                      [1]     {Section~\ref{#1}}

\usepackage{changes} 
\definechangesauthor[name={Rohit}, color=blue]{A}

\title{\LARGE \bf Evolutionary Continuous Adaptive RL-Powered Co-Design for Humanoid Chin-Up Performance}

\author{Tianyi Jin$^{1,4*}$, Melya Boukheddimi$^{1*}$, Rohit Kumar$^{1}$, Gabriele Fadini$^{2}$, and Frank Kirchner$^{1,3}$
\thanks{This work was supported by the CoEx (FKZ 24883) project funded by the German Aerospace Center (DLR) with	federal funds from the Federal Ministry of Education and Research (BMBF).}
\thanks{{$^{*}$Both authors contributed equally to the paper}}
\thanks{$^{1}$Robotics Innovation Center, DFKI GmbH, 28359 Bremen, Germany.}%
\thanks{$^{2}$Computational Robotics Lab, ETH Zurich.}%
\thanks{$^{3}$AG Robotik University of Bremen, 28359 Bremen, Germany.}%
\thanks{$^{4}$RWTH Aachen University, Aachen, Germany.}%
\thanks{{Corresponding author}:
   \href{mailto:sun\_ner@outlook.com}{sun\_ner@outlook.com}}   
}

\begin{document}
\newcommand{\robotName}{RH5V2}

\newcommand{\mvec}[1]{\bm{#1}}
\newcommand{\vc}[1]{\mathbf{\mathbf{#1}}}

\newcommand{\q}{\textbf{q}}
\newcommand{\dq}{\dot{\q}}
\newcommand{\ddq}{\ddot{\q}}


\newcommand{\Mass}{\mathbf{M}}
\newcommand{\Bias}{\mathbf{b}}
\newcommand{\Gravity}{\mathbf{g}}
\newcommand{\Force}{\mathbf{\lambda}}
\newcommand{\Torque}{\mathbf{\tau}}
\newcommand{\Jac}{\mathbf{J}}

\newcommand{\BIN}{\begin{bmatrix}}
\newcommand{\BOUT}{\end{bmatrix}}

\newcommand{\sref}[1]{Sec~\ref{#1}}
\newcommand{\eref}[1]{(\ref{#1})}
\newcommand{\fref}[1]{Fig.~\ref{#1}}
\newcommand{\tref}[1]{Table~\ref{#1}}
\newcommand{\equationref}[1]{Equ.~\ref{#1}}
\newcommand{\state}{\mathbf{x}}
\newcommand{\ctrl}{\mathbf{u}}
\newcommand{\dynsys}{\mathbf{f}}

\newcommand{\qTr}{\underline{\q}}
\newcommand{\dqTr}{\underline{\dq}}
\newcommand{\ddqTr}{\underline{\ddq}}
\newcommand{\TorqueTr}{\underline{\Torque}}

\newcommand{\costl}{l}
\newcommand{\dts}{\Delta t_s}
\newcommand{\st}{\text{subject to}}

\maketitle
\thispagestyle{empty}
\pagestyle{empty}
\begin{abstract}
Humanoid robots have seen significant advancements in both design and control, with a growing emphasis on integrating these aspects to enhance overall performance. Traditionally, robot design has followed a sequential process, where control algorithms are developed after the hardware is finalized. However, this can be myopic and prevent robots to fully exploit their hardware capabilities. Recent approaches advocate for co-design, optimizing both design and control in parallel to maximize robotic capabilities.
This paper presents the Evolutionary Continuous Adaptive RL-based Co-Design (\mbox{EA-CoRL}) framework, which combines reinforcement learning (RL) with evolutionary strategies to enable continuous adaptation of the control policy to the hardware. \mbox{EA-CoRL} comprises two key components: Design Evolution, which explores the hardware choices using an evolutionary algorithm to identify efficient configurations, and Policy Continuous Adaptation, which fine-tunes a task-specific control policy across evolving designs to maximize performance rewards.
We evaluate \mbox{EA-CoRL} by co-designing the actuators (gear ratios) and control policy of the RH5 humanoid for a highly dynamic chin-up task, previously unfeasible due to actuator limitations. Comparative results against state-of-the-art RL-based co-design methods show that \mbox{EA-CoRL} achieves higher fitness score and broader design space exploration, highlighting the critical role of continuous policy adaptation in robot co-design.
\end{abstract}
\section{Introduction}
\label{sec:intro}
The field of humanoid robotics is evolving rapidly, yet robot design still faces significant challenges, especially for demanding tasks. These include lifting heavy loads, executing fast and repetitive assembly operations, supporting physical rehabilitation, and mastering highly dynamic movements such as jumping, back-flips, and parkour.
Relevant literature shows active but also limited humanoid performance for laborious tasks. Past work has dealt with the design of robots capable of heavy-duty operations.
\cite{dallali2013designing} simulates a set of rescue-relevant tasks on the humanoid COMAN to evaluate joint compliance, but is limited by simplified model and idealized dynamics in the simulator.
\cite{abi2019torque} presents DLR humanoid TORO pushing a 50kg table based on a whole-body balancing controller, but ignores the real robot actuators' limit. 
\begin{figure}[h]
	\centering
	\includegraphics[width =0.5\textwidth, height=7.5cm]{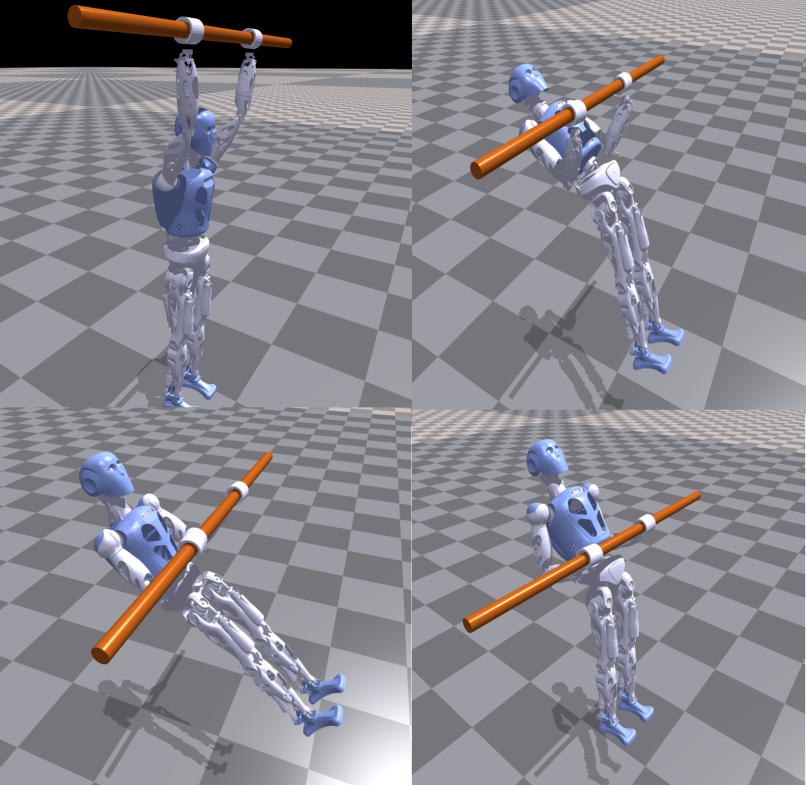}
	\caption{Chin-up RL policies on the RH5 humanoid designs obtained through \mbox{EA-CoRL}.}
	\label{fig:chinUp}
\end{figure}
One of the latest results, \cite{falcon}, introduces a framework that combines lower and upper body reinforcement learning (RL) agents to achieve robust humanoid loco-manipulation tasks with adaptive force control, while respecting actuator limits. While the work is impressive, demonstrating highly dynamic capabilities such as an 82 kg cart-pulling motion, it is limited to improving control only after the robot has been built. The design of the robot model itself is not addressed.
\cite{Polverini2020} proposes a push control strategy on a 120kg cabinet for the CENTAURO robot, yet the approach relies on its centaur-type leg design contacting the wall as support. 
\cite{HumanoidBench2024} utilizes the Unitree H1 model to simulate a benchmark of whole-body manipulation tasks using RL. However, the overall performance presented in this work for high-effort tasks, such as heavy package moving and power lifting, remains very sub-optimal with low task rewards.
Traditionally, robotic design has followed a sequential approach: first, the system is built, then tested, and subsequently refined through multiple iterations. This process designing, building, testing, and re-designing relies heavily on the intuition and expertise of mechanical designers, with the shortcomings of being often sub-optimal.
Recent research has increasingly focused on co-design methodologies. The aim is to integrate control and hardware in the design process, enabling a synergistic approach where mechanical and control aspects are co-optimized. This approach seeks to optimize the robot's structure for its intended goals, behaviors, or motion capabilities.
\paragraph{Advances in Co-Design Approaches}
Several co-design strategies have emerged in recent years. One of the earliest approaches involves optimal control (OC) based co-design strategies. For instance, \cite{fadini2024co} presents a framework where the design is optimized to achieve dynamic motion while minimizing energy consumption in a quadruped robot. 
In \cite{sartore2023}, a co-design approach based on OC is developed to achieve an optimized design guided by ergonomic criteria during interactions with various agents. Simulation results using humanoid models demonstrate a reduction in energy consumption.
Another approach is the bi-level kineto-static formulation of co-design used in \cite{10842359}, which optimizes a generic manipulator to enhance its range of motion.
\cite{kumar2025parallel} focuses on the co-design of a parallel belt-driven manipulator, incorporating the constraints of the belt's parallel coupling into the robot's dynamics and optimizing its transmission ratios. 
Additionally, \cite{8078267} focuses on the design optimization of a robotic prosthesis based on its variable stiffness actuators (VSA), demonstrating improved performance and highlighting the potential of co-design applications in VSA-based systems.
Although these methods achieve their intended objectives, they exhibit notable limitations. They are extremely dependent on the underlying model and their applicability falls short in the case of environmental disturbances (e.g., damping, friction, sensor noise and delays), and system uncertainties. This lack of robustness limits their applicability to real-world scenarios and their ability to generalize to model or environment variations. Consequently, the motions generated through OC remain highly dependent on the underlying models, making them vulnerable to deviations from idealized conditions.
\paragraph{Reinforcement Learning-Based Co-Design}
Reinforcement learning presents a promising alternative by offering a model-free control approach that enhances adaptability and robustness. In co-design, RL leverages domain randomization \cite{Ibarz2021}, to expose policies to diverse conditions, improving their resilience to model inaccuracies and environmental uncertainties. This could make RL-based co-design potentially more robust and suitable for real-world deployment compared to methods based on OC \cite{MetaLoco}. However, RL policies are generally closely linked to the design parameters with which they were trained, requiring adaptation or retraining when the design changes. This coupling of policy and design highlights the need for methods that can generalize over design variations by efficiently  adapting policies to new designs.
A notable contribution in the domain of RL-based co-design is \cite{METARL}, which introduces a model-free meta-RL co-design framework implemented on a quadruped locomotion task. This approach successfully adapts to random velocity commands on diverse rough terrains and has been validated on real-world systems. 
In \cite{bjelonic-learning}, the authors co-optimize a parallel elastic joint for quadrupedal robot locomotion. 
Similarly, \cite{chen2024} addresses highly dynamic quadruped parkour motion by proposing a novel pre-training fine-tuning co-design algorithm that ensures time efficiency while integrating optimization strategies for different robotic configurations. 
Furthermore, \cite{cheng2024structuralopt} combines RL with evolutionary algorithms to achieve structural co-design for a lightweight bipedal robot, optimizing its design specifically for the gait.
These advances highlight the growing potential of RL-driven co-design, offering novel pathways for developing robotic co-design platforms beyond conventional techniques \cite{fadini2024making}. 
While previous works have predominantly focused on optimizing link lengths within robotic systems, the design space remains vast, requiring more comprehensive studies that consider various other design aspects. 
Another aspect is to focus on developing more versatile and generalized learning-based co-design frameworks that can be applied to a broader range of robotic systems, ensuring minimal modifications to existing architectures.
\paragraph*{Contribution}
In this work, we propose a generic formalization of an Evolutionary Continuous Adaptive RL-based Co-Design algorithm, referred to in this paper as "\mbox{EA-CoRL}". This framework enables RL-based robotic co-design and is validated through a case-study of a whole-body humanoid robot executing a highly dynamic chin-up motion.
The main contributions of this work are as follows:
\begin{itemize}
\item A novel RL-based framework (\mbox{EA-CoRL}) for the co-design of both robot model and control policies.
\item \mbox{EA-CoRL} integrates a continuous adaptive co-design process that enables broader exploration of the design space, while enhancing performance consistency and reducing the chances of premature convergence compared to a baseline approach.
\item Development of a model-agnostic humanoid chin-up policy, used as a case study to validate the co-design capabilities for whole-body humanoid motion.
\item Demonstration that a high-effort task, previously unfeasible due to actuator limitations, can be achieved without significant hardware modifications, through co-design of motor gear ratios and a RL policy optimized on hardware. This approach broadens the scope of co-design while minimizing hardware costs.
\end{itemize}
\paragraph*{Organization}
\rsec{sec:methodology} presents the methodology of the Evolutionary Continuous Adaptive RL-based Co-Design approach.
\rsec{sec:implementation}  describes the implementation details of the experimental setup used to evaluate the approach.
\rsec{sec:results} presents the experimental results along with a comparative analysis and discussion. 
Finally, \rsec{sec:conclusion} summarizes the key findings and outlines potential future research directions.
\section{Methodology}
\label{sec:methodology} 
\begin{figure*}[!htbp]
	\centering
	\includegraphics[height=5.5cm]{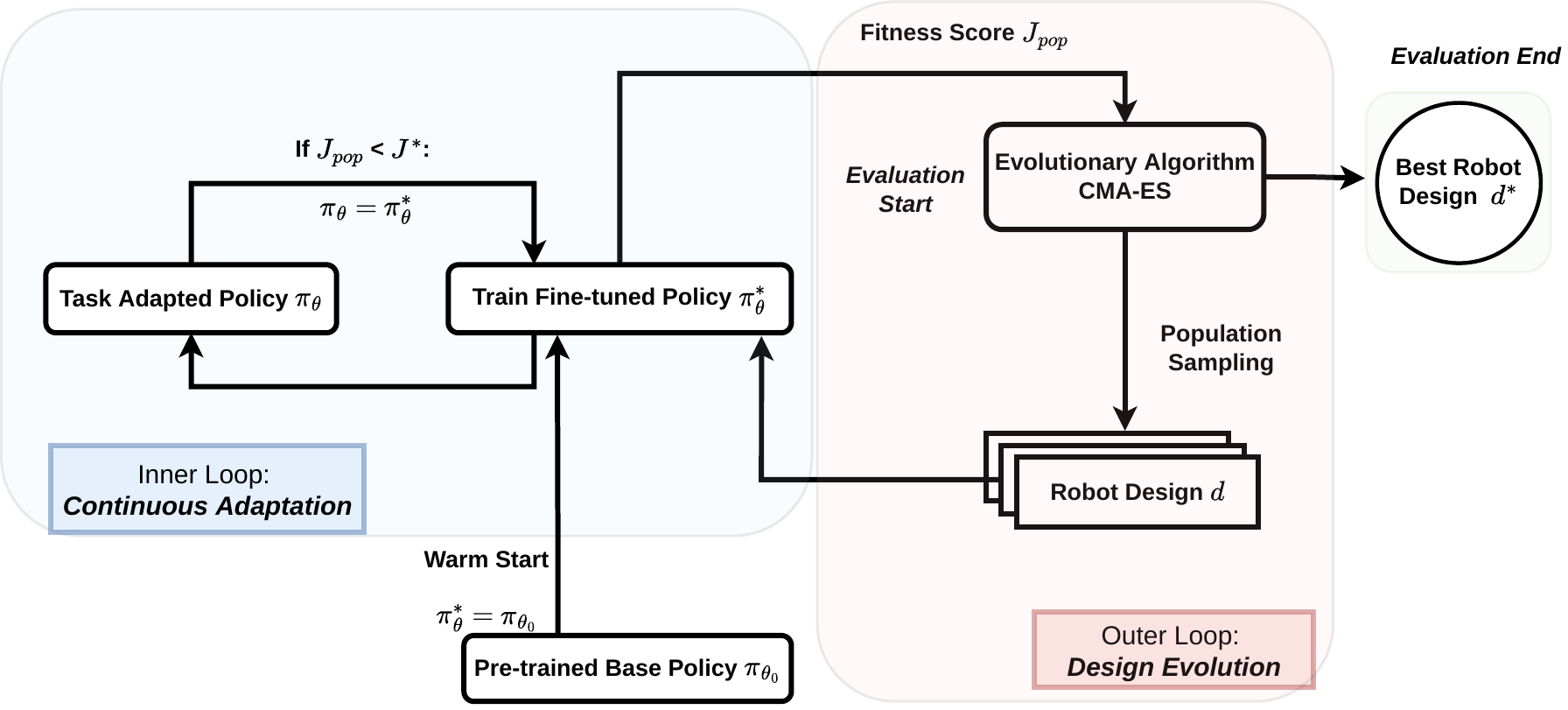}  
	\caption{Overview of the \mbox{EA-CoRL} framework methodology.}
	\label{fig:pipeline}
\end{figure*} 
In this section, we introduce \mbox{EA-CoRL}, an Evolutionary Continuous Adaptive RL-based Co-Design that synergistically optimizes both robot design and control policy through a combination of an evolutionary strategy and RL.
\begin{itemize}
\item \textbf{Evolutionary} refers to the iterative exploration of robot designs using an evolutionary algorithm, enabling convergence to best robot configurations.
\item \textbf{Continuous Adaptive} denotes the ongoing fine-tuning of the control policy across new design candidates at each evolutionary step, ensuring the policy is never frozen but dynamically updated to maximize the Fine-tuned Policy.
\end{itemize}
\fref{fig:pipeline} illustrates the structure of \mbox{EA-CoRL}, which consists of two interconnected components: Design Evolution and policy Continuous Adaptation.
\subsection{Design Evolution: Outer Loop (Red Block)}
\label{sec:methodology:Design}
This phase explores design parameters to achieve robot co-design. To efficiently navigate the {space of design parameters \( \mathcal{D}\)}, we employ the Covariance Matrix Adaptation Evolution Strategy (CMA-ES) \cite{cmae}, a widely used evolutionary algorithm for black-box optimization in robot design search.
In each {CMA-ES} iteration, the algorithm iteratively samples candidates designs from a multivariate Gaussian distribution within {\( \mathcal{D}\)}. It then evaluates their performances based on a {Fitness Score \(\mathcal{J}_{pop}\)}, derived from the {RL training results of each candidate design \(d \)}. 
The highest-performing candidates are selected and evolved into the next generation, progressively refining the designs toward a maximized solution. After completing the {design evolution} iterations (outer loop), the best-performing robot designs \(d^* \) are selected based on their 
maximized policy {reward \(\mathcal{R} \)} i.e.
minimized {Fitness Score \(\mathcal{J^*}\)} (Best Fitness Score).
\subsection{Continuous Adaptation: Inner Loop (Blue Block)} 
In this phase, {parallel simulation environments} are instantiated in a dynamic simulator {NVIDIA Isaac Gym} \cite{2021isaacgym} based on {candidates designs} generated by the {Design Evolution} phase.
This phase consists of multiple components that enable efficient policy continuous adaptation: 
\subsubsection{Base Policy \(\pi_{\theta_0} \)} 
is pre-trained on the initial robot designs \(\mathcal{D}_0\) generated by the {Design Evolution} phase.
Leveraging a base policy provides a warm start for {Fine-tuned Policy \( \pi_{\theta^*} \) }, which is used to fine-tune robot configurations generated by CMA-ES.
Due to the varying design candidates in the co-design process, directly deploying a base policy to evaluate all robot models may lead to low task reward, while fine-tuning each individual candidate may cause huge computational costs during evaluation.
To address these issues, we introduce the {Fine-tuned Policy \( \pi_{\theta^*} \)} .
\subsubsection{Fine-tuned Policy \( \pi_{\theta^*} \)}  
is initialized at the first iteration of the \textit{Design Evolution} (outer loop) with the Base Policy, i.e., \(\pi_{\theta_0} \) = \( \pi_{\theta^*} \).
This policy is then fine-tuned with respect to each new robot design's population \(\mathcal{D}_i\) at each iteration \(i\) of the \textit{Design Evolution} loop.
The obtained reward from training this fine-tuned policy is then used to evaluate the designs \(\mathcal{D}_i\).
\textbf{If} the condition, \( J_{\text{pop}} < \mathcal{J}^* \) is satisfied, then for the next iteration of the \textit{Design Evolution}, \( \pi_{\theta^*} \) is updated with the obtained fine-tuned policy, referred to here as the task-adapted policy \( \pi_{\theta} \).
\subsubsection{Task-Adapted Policy \( \pi_{\theta} \)}  
is derived from the fine-tuned policy \( \pi_{\theta^*} \) at each subsequent iteration of the \textit{Design Evolution}, following the fine-tuning stage.  
If the task-adapted policy \( \pi_{\theta} \) achieves a higher reward, the fine-tuned policy \( \pi_{\theta^*} \) is updated accordingly.  
This dynamic policy update mechanism enables exploration of maximized motion performance across diverse robot configurations, ultimately leading to a refined final design \( d^* \) along with an optimized policy \( \pi_{\theta^*} \).
\subsubsection{RL Training}
our approach trains a design-aware policy \( \pi_{\theta^*} \) capable of adapting to various robot designs within the design space \(\mathcal{D}\). The problem is modeled as a Markov Decision Process, defined by \((S, A, P, r, \gamma)\). At each time step \(t\), the agent observes the current state \(s_t \in S\) from the environment, selects an action \(a_t \in A\), transfers to a new state \(s_{t+1}\) based on transition probabilities \(P(s_{t+1} \mid s_t, a_t)\), and receives a reward \(r_t \in r\). The objective of the {RL Training} is to derive the best policy  \( \pi_{\theta^*} \) by maximizing the expected cumulative discounted return, characterized by the state-value function \eqref{eq:optimal_value}, where \(\gamma\) serves as a discount factor:
\begin{equation}
	V^*(s) = \max_{\pi_\theta} \mathbb{E}\left[ \sum_{t=0}^{\infty} \gamma^t r_t \Big|s_t = s \right] \label{eq:optimal_value} 
\end{equation}
To solve this problem, we adopt the model-free RL algorithm Proximal Policy Optimization (PPO) \cite{schulman2017proximal}, which has demonstrated performance in robotic applications, including legged locomotion \cite{rudin2022learning}. By training samples of {candidates designs} in parallel, the policy learns to generalize throughout the design space \(\mathcal{D} \), allowing it to adapt effectively to different robot configurations.
\subsection{Best Robot Design \(d^* \):} 
After completing the {design evolution} iterations (outer loop), the best-performing robot design variables are selected based on their maximized policy {reward \(\mathcal{R} \)} i.e.  minimized {fitness score \(\mathcal{J^*}\)}, throughout the {design evolution} process.
\section{EXPERIMENTAL SETUP}
\label{sec:implementation} 
In this section, we evaluate our co-design method through an experiment focused on the chin-up task applied to the humanoid robot RH5 \cite{boukheddimi2022introducing}. 
First, we introduce the RH5 robot and the chin-up task. Then, we describe the implementation details of the \mbox{EA-CoRL} approach.
\subsection{RH5 Whole-Body Humanoid}
The goal is	 to co-design the third-generation whole-body humanoid RH5, as shown in \fref{fig:chinUp}, to achieve a chin-up task. RH5 weighs 55 kg and stands 187 cm tall. It integrates the upper-body RH5v2, introduced as a high-performance humanoid in \cite{boukheddimi2022introducing}, \cite{MelyaLift}, with the lower limbs of RH5v1, designed for fast walking, as detailed in \cite{bergonzani2023fast}.
These previous works already demonstrate the highly dynamic capabilities of RH5, a state-of-the-art humanoid featuring a series-parallel hybrid design and 33 degrees of freedom (DoFs). 
Despite these advanced features, RH5 encounters challenges in executing high-force tasks e.g.\  chin-ups due to actuator limitations. To overcome these challenges, this study utilizes the proposed \mbox{EA-CoRL} methodology, which co-design the gear ratios of RH5's actuators, enhancing its ability to perform dynamic chin-up efficiently.
\subsection{Chin-Up: Evaluating High-Force, Dynamic Performance}
The chin-up exercise \cite{youdas2010surface} is the selected study-task for humanoid robot to assess whole-body strength and evaluate demanding, high-force tasks performance. Humanoid chin-up involves lifting the body until the feet leave the floor and the head rises above the bar.
For the original RH5 model, this motion is not achievable while respecting its joint limits, thus it represents a challenge for its actuation system. 
\subsection{\mbox{EA-CoRL} Experimental Details for RH5's Chin-Up Task}
\subsubsection{Design Space}
As the collection of design parameters, specifically the gear ratios, which characterizes the actuation properties of the robot. Each actuator design of the RH5 is parameterized by a vector of gear ratio factors: \( d = [d_{\text{leg}}, d_{\text{shoulder}}, d_{\text{elbow}}, d_{\text{wrist}}] \), where each element represents a gear ratio factor affecting specific joint groups. These groups are structured as follows:
\begin{itemize}
	\item Lower limb group (\( d_{\text{leg}} \), 5 DoFs): body pitch (1), hip flexion-extension (2: right, left), knee (2: right, left).
	\item Shoulder group (\( d_{\text{shoulder}} \), 6 DoFs): three rotations (right, left).
	\item Elbow group (\( d_{\text{elbow}} \), 2 DoFs): flexion-extension (right, left).
	\item Wrist group (\( d_{\text{wrist}} \), 4 DoFs): pitch and yaw (right, left).
\end{itemize}
Each gear ratio factor is bounded within predefined limits during evolution phase. The candidate designs are sampled via CMA-ES algorithm and evaluated in the NVIDIA Isaac Gym dynamic simulator \cite{2021isaacgym} within highly {Parallel Simulation Environments}, as described in \sref{sec:methodology:Design}.
To model realistic actuators, the considered parameter $d$ directly affects the robot’s joint torque and velocity limits, denoted by $\tau_{\max}$ and $\dot{q}_{\max}$, as expressed in \eqref{eq:torqurLimit} and \eqref{eq:veloLimit}:  
\begin{equation}
\label{eq:torqurLimit}
\tau_{\max}(d) = \tau_{\text{default}} \times d
\end{equation}
\begin{equation}
\label{eq:veloLimit}
\dot{q}_{\max}(d) = \frac{\dot{q}_{\text{default}}}{d}
\end{equation}
Here, $\tau_{\text{default}}$ and $\dot{q}_{\text{default}}$ represent the original robot’s joint torque and velocity limits, respectively.
\subsubsection{Observation Space}
The agent's observation space \(o_t\) integrates standard proprioceptive data \(\hat{o}_t\) and  privileged gear ratio factors information \(o_t^\text{gr}\) to enhance policy fine-tuning across varying robot designs. The proprioceptive measurement \(\hat{o}_t\) comprises features such as target position distance, command, gravity projection, joint positions, joint velocities, and actions. 
Additionally, the gear ratio factors serve as privileged information within the observation space, providing additional context for the actor-critic network during training, thereby facilitating policy fine-tuning to different robot configurations.  
The gear ratio factors are encoded into a latent space and then stacked with proprioceptive data. 
\subsubsection{Action Space}
The action space \(a_t\) consists of continuous position and velocity commands for 17 of the 33 DoFs of RH5, selected to balance training simplicity with essential movement capabilities.  This subset includes the 6 actuated DoFs in each arm, the body pitch joint (1 DoF), and the knee and hip flexion-extension joints in each leg. Position control is implemented via a PD controller, which tracks target joint positions and velocities and computes the necessary torques to ensure smooth and precise movements, thereby maximizing task performance.
\subsubsection{Reward Function}
The reward function \(R_t\) is designed and tuned to balance task performance, motion regularization, and energy efficiency. A detailed description of the terms and hyper-parameters are provided in \tref{tab:rh5-rewards}.
\begin{equation}
	R_t = \sum_{i=1}^{n} w_i\,r_i(s_t, a_t)
	\label{eq:rewardFunction}
\end{equation}
Equation \eqref{eq:rewardFunction} formulates the reward function as a weighted sum of various reward terms, including chin-up-specific terms that encourage successful motion behavior. 
In addition, regularization penalties mitigate joint limit overshoot, abrupt control changes, excessive torque output, and asymmetry motion between the left and right sides. 
\begin{table*}[ht!]
	\centering
	\caption{Multi-term Reward Functions Definition for the RH5 Chin-Up Task}
	\label{tab:rh5-rewards}
	\begin{tabular}{l l c}
		\\[-15pt]
		\midrule
		\textbf{Reward Term} & \textbf{Mathematical Expression} & \textbf{Weights}\\
		\midrule
		\texttt{Chinup} &
		\(
		r_{\text{chinup}} = \exp\left(- \left\| \text{pos}_{\text{head}} - \text{pos}_{\text{goal}} \right\|^2 \right)
		\) & 30.0
		\\[3pt]
		\texttt{Hollow Cylinder} &
		\(
		r_{\mathrm{hCyl}}
		= \begin{cases}
			0, & 0.5 < (y_{cyl_L} - y_{cyl_R}) < 0.8, \\
			10, & \text{otherwise}.
		\end{cases}
		\) & -2.0
		\\[3pt]
		\texttt{Base Position} &
		\(
		r_{\mathrm{base}}
		= \begin{cases}
			0, & if  \,
			x_{\text{base}} < x_{\text{highbar}} \, or \, \bigl(\tfrac{z_{feet_L} + z_{feet_R}}{2} \bigr) < z_{\text{base}} < z_{\text{highbar}}\\
			20,   & \text{otherwise}.
		\end{cases}
		\) & -2.0
		\\[3pt]
		\texttt{Joint Regularization} &
		\(
		r_{\mathrm{jReg}}
		= \sum_{(i,j)\in \mathcal{J}}
		\exp\!\Bigl(-\,\bigl (q_i - q_j \bigr)^{2}\Bigr),
		\)
		where \(\mathcal{J}\), a set of symmetric DoFs. & -5.0
		\\[3pt]
		\texttt{Orientation} &
		\(
		r_{\mathrm{ori}}
		= \|\mathbf{g}_{\mathrm{proj},xy}\|^{2}.
		\) & -5.0
		\\[3pt]
		\texttt{Torque} &
		\(
		r_{\mathrm{\tau}}
		= \sum_{i} \tau_{i}^{2}.
		\) & $-1 \cdot 10^{-5}$
		
		\\[3pt]
		\texttt{Joint Acceleration} &
		\(
		r_{\mathrm{acc}}
		= \bigl\|\tfrac{\dot{\mathbf{q}}_{t} 
			- \dot{\mathbf{q}}_{t-1}}{\Delta t}\bigr\|^{2}.
		\) & $-1 \cdot 10^{-5}$
		\\[3pt]
		\texttt{Action Rate} &
		\(
		r_{\mathrm{act}}
		= \|\mathbf{a}_{t} - \mathbf{a}_{t-1}\|^{2}.
		\) & $-1 \cdot 10^{-3}$
		\\[3pt]
		\texttt{Joint Position Limit} &
		\(
		r_{\mathrm{pos_{lim}}}
		= \sum_{i} \Bigl[\max\{0,\,q_{\min,i} - q_i\}
		+ \max\{0,\,q_i - q_{\max,i}\}\Bigr].
		\) & -2.0
		\\[3pt]
		\texttt{Joint Velocity Limit} &
		\(
		r_{\mathrm{v_{lim}}}
		= \sum_{i}\;\text{clip}\bigl(|\dot{q}_i| - \dot{q}_{i,\max},\,0,\,1\bigr).
		\) & -2.0
		\\[3pt]
		\texttt{Joint Torque Limit} &
		\(
		r_{\mathrm{\tau_{lim}}}
		= \sum_{i}\;\text{clip}\bigl(|\tau_i| - \tau_{i,\max},\,0,\,1\bigr).
		\) & -2.0
		\\
		\bottomrule
	\end{tabular}
\end{table*}
\subsection{\mbox{EA-CoRL} Deployment Workflow}
\label{subsec:implementation}
\begin{algorithm}[htb]
	\caption{\mbox{EA-CoRL} co-design pseudo-code}
	\label{alg:cmaes_rl}
	\SetKwInOut{Input}{Input}
	\SetKwInOut{Output}{Output}
	\SetKwInOut{Init}{Init}
	\Input{
	    Number of populations \( N_{\text{pop}} \)\\
	    Number of training environments \( N_{\text{env}} \)\\
	    CMA-ES hyperparameters: \( \theta_0, \sigma_0, \mathcal{O}_{\text{CMA-ES}} \)
	}
	\Output{
	    Best: [design \( d^* \), fitness score \( \mathcal{J}^* \), Fine-tuned Policy \( \pi_{\theta^*} \)]
	}
	\Init{
	    CMA-ES with \( \theta_0, \sigma_0, \mathcal{O}_{\text{CMA-ES}} \)\\
	    \( N_{\text{exp}} =  \frac{N_{\text{env}}}{N_{\text{pop}}} \),
	    best fitness score \( \mathcal{J}^* \)
	}
	\For{$i_e \in [1, N_{\text{designs\_evolution}}]$}{
	    Generate candidate design solutions \( \mathcal{D}' \)\\
	    Expand \( \mathcal{D}' \) to parallel training environments\\
	    
	    \If{$i_e == 1$}{  
	        Pre-train the Base Policy \(\pi_{\theta_0}\) as warm-start\\
	        \( \pi_{\theta^*} \gets \pi_{\theta_0} \),
	        \( \mathcal{J}^* \gets J_{\text{pop},1} \)
	    }
	    \Else{
	        Train the Task Adapted Policy \(\pi_{\theta_i} \) from the Fine-tuned Policy \( \pi_{\theta^*}\)
	        Compute fitness score \( J_{\text{pop}, i} \)\\ 
	        \If{\( J_{\text{pop}, i} < \mathcal{J}^* \)}{
	            Continuous Policy Adaptation: 
	            \( \pi_{\theta^*} \gets \pi_{\theta_i} \)\\
	            Update best fitness score: \( \mathcal{J}^* \gets J_{\text{pop}, i} \)
	        } 
	    }
	    Set best design: \( d^* \gets d_{\mathcal{J}^*} \) \\
	    Evolve CMA-ES   
	}
	\Return{\( d^* \), \( \mathcal{J}^* \), \( \pi_{\theta^*} \)}
\end{algorithm}
The pseudo-code in Algorithm~\ref{alg:cmaes_rl} explicitly outlines the deployment workflow of the \mbox{EA-CoRL} method.
The \mbox{EA-CoRL} initializes CMA-ES with maximization options \( \mathcal{O}_{\text{CMA-ES}} \), using the hyper-parameters listed in \tref{tab:hyperparams}. During each CMA-ES iteration, the four gear ratio factor designs \( d_{\text{leg}}, d_{\text{shoulder}}, d_{\text{elbow}}, d_{\text{wrist}} \) are sampled and expanded  to match the number of parallel training environments in the simulator \cite{2021isaacgym}. The matching process is defined as in \eqref{equ:Deployment}:
\begin{equation}
	\label{equ:Deployment}
	d_{\text{expand}} = \{ d_k \}_{k=1}^{N_{\text{env}}}
\end{equation}
\begin{equation*}
	\text{where}  
	\quad d_k = d_j,
	\quad j = \left\lceil \frac{k}{N_{\text{exp}}} \right\rceil,
	\quad N_{\text{exp}} = \frac{N_{\text{env}}}{N_{\text{pop}}} 
\end{equation*}
\( d_k \) denote the gear ratio factors design for environment $k$ under simulation. 
\( d_{expand} \) is the entire set of gear ratio factor designs applied to all environments under simulation.
\( d_j \) represents a sampled designs from the design space.
\( N_{\text{pop}} \) is the number of candidate designs generated by CMA-ES.
\( N_{\text{exp}} \) defines the number of environments assigned per design.
This allocation ensures that each candidate design undergoes evaluation over multiple rollouts, mitigating reward noise and enhancing robustness in objective function computation.
A total of 50 populations of RH5 robot models are instantiated, incorporating the expanded solutions \( d_{\text{expand}} \) to set joint effort and velocity limits across 4000 parallel environments. Each population is evaluated over 80 simulation rollouts. 
A base policy \( \pi_0 \) is pre-trained on the initial sampling set \( \mathcal{D}_0 \) for 5000 iterations to provide a warm start for fine-tuning. The Fitness Score $J_{\text{pop}}$ is defined as the negative mean reward $R$ per population, averaged across all expanded environments:
\begin{equation}
	\label{equ:objective}
	J_{\text{pop}} = \frac{1}{N_{\text{exp}}} \sum_{i=1}^{N_{\text{exp}}} -R(d_{\text{expand}})
\end{equation}
\begin{equation}
	\label{equ:BestDes}
	d^* = \arg\min_d J_{\text{pop}}(d)
\end{equation}
Based on \eqref{equ:objective}, the reward is naturally inversely proportional to the fitness score. In the results section, the lowest fitness score will always correspond to the highest reward value, representing the best gear ratio factors for the best design \eqref{equ:BestDes}.
Following the CMA-ES evaluation, the 10 lowest fitness scores \( J_{\text{pop}} \) are selected to guide the next evolutionary update. If a new best score \( \mathcal{J}^* \) is found, the Best RL policy \( \pi_{\theta^*} \) is updated accordingly. The process continues until CMA-ES reaches the final iteration of the outer loop and returns the Best Robot Design \(d^*\) and Best policy \(\pi_{\theta^*}\).
The \mbox{EA-CoRL} algorithm is executed on a single NVIDIA RTX A6000 GPU, requiring approximately four days to complete the full co-design process and identify the {Best Robot Design \(d^* \)}. 
In this implementation, the CMA-ES algorithm is implemented using PyCMA library \cite{hansen10cma}.
\begin{table}[htb]
	\centering
	\caption{\mbox{EA-CoRL} Hyper-parameters}
	\label{tab:hyperparams}
	\begin{tabular}{|l|c|}
		\\[-15pt]
		\hline 
		\textbf{Hyper-parameter} & \textbf{Value} \\
		\hline
		Population Size ($N_{\text{pop}}$) & 50 \\
		Max Evolution Iterations ($N_{\text{evol}}$) & 50 \\
		Initial Mean ($\theta_0$) & 0.2 \\
		Initial Standard Deviation ($\sigma_0$) & 0.3 \\
		Mutation Rate ($\text{CMA}_{\text{mu}}$) & 10 \\
		Design Parameter Boundaries & [0.5 , 4.0] \\
		Number of Training Environments ($N_{\text{env}}$) & 4000 \\
		Policy Adaptation Training Steps ($N_{\text{train}}$) & 2500 \\
		Base Policy Training Steps ($N_{\text{train}}$) & 5000 \\
		Adaptive RL Learning Rate ($\alpha$) & $1 \times e^{-5}$ \\
		\hline
	\end{tabular}
\end{table}
\section{Results and Discussion}
\label{sec:results} 
This section presents and analyzes the experimental results of the RH5 co-design for the chin-up task using the proposed \mbox{EA-CoRL} method. 
Before presenting the results, we define the baseline approach used for comparison.

\paragraph*{PreTraining-FineTuning (PT-FT)}
We use the acronym PT-FT for "PreTraining-FineTuning." This co-design algorithm was successfully applied in \cite{METARL} to optimize legged robots. 
Unlike \mbox{EA-CoRL}, PT-FT begins by pre-training a universal policy that covers all robot designs in the design space using meta-RL, followed by rapid fine-tuning for specific design candidates generated by an evolutionary algorithm. 
The main difference lies in how the policy is updated: PT-FT freezes the fine-tuned policy, while \mbox{EA-CoRL} continuously refines it through Continuous Adaptation.

The remainder of this section is structured into three parts. 
First, we present the best robot designs and analyze correlations between design parameters across the design space. 
We evaluate the performance of \mbox{EA-CoRL} by comparing it with PT-FT. 
Secondly, we discuss the obtained best gear ratios. 
Lastly, the section discusses the proposed approach and its results.

\subsection{Results}
The results are generated for six independent runs with different random seeds. 

\begin{figure}[t]
	\centering
	\includegraphics[height=7cm]{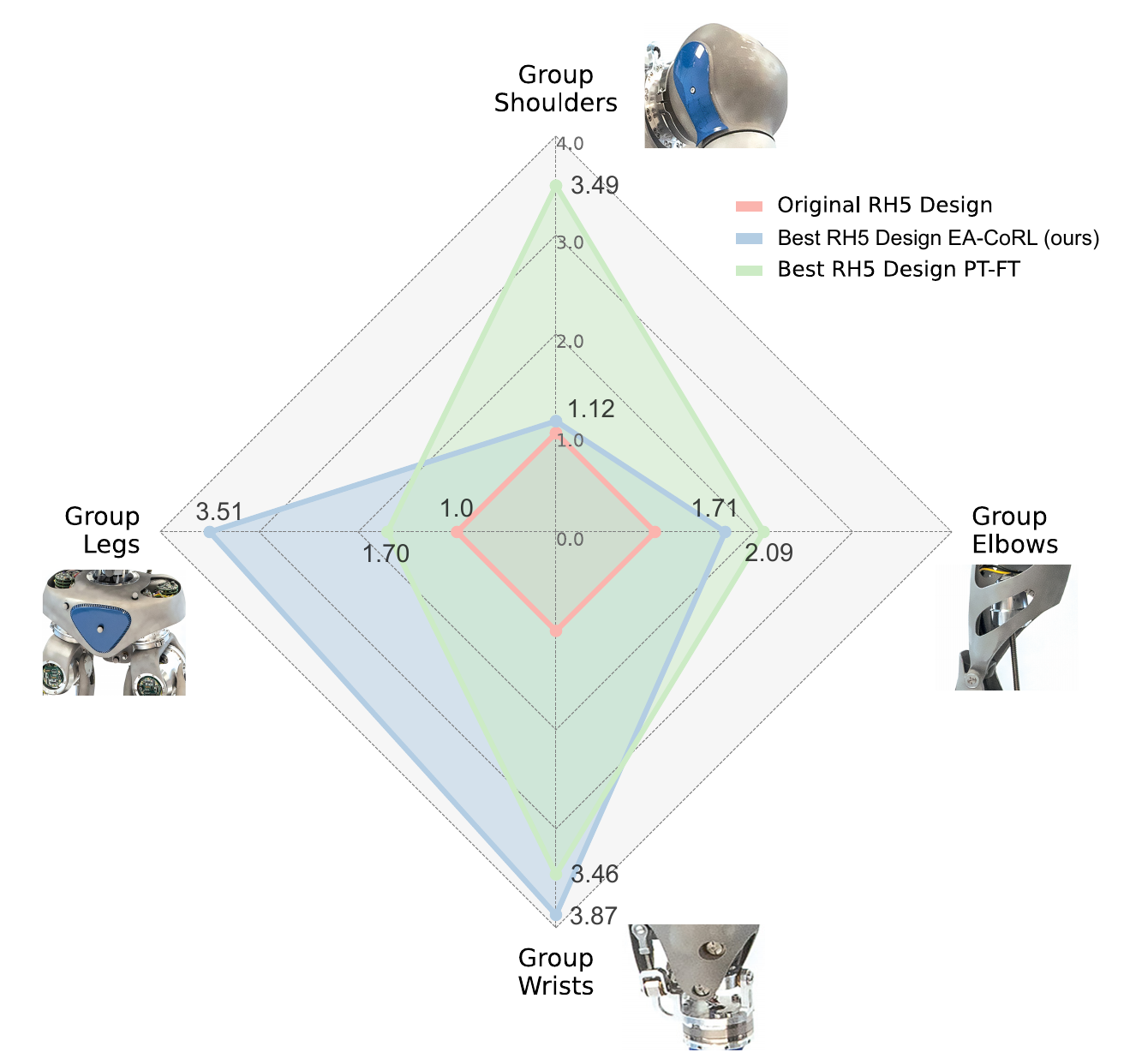}
	\caption{Original and obtained best RH5 gear ratio factors for the chin-up task using \mbox{EA-CoRL} (ours) and PT-FT methods.}
	\label{fig:radar_chart}
\end{figure}
Figure \ref{fig:radar_chart} presents the original and best RH5 gear-ratio factor designs obtained using the \mbox{EA-CoRL} and PT-FT methods, respectively. 
Results from \mbox{EA-CoRL} show that the leg and wrist groups require higher gear ratio factors than the shoulder and elbow groups to successfully perform the chin-up motion. 
Both the leg and wrist joint groups actively contribute to the chin-up motion, playing a key role in maintaining the robot's balance, as observed in the attached video. 
In the original RH5 design, the wrist actuators are capable of generating only one-fifth of the maximum torque output of the shoulder actuators, which justifies the need for increased gear ratio factors in both the wrist and leg groups. 
In comparison to \mbox{EA-CoRL}, the best gear-ratio factor design from PT-FT features lower gear ratio factors for the leg group but higher ratios for the shoulder, elbow, and wrist groups. 
This trade-off highlights the flexibility of the \mbox{EA-CoRL} framework and the effectiveness of the co-design approach, where the four gear-ratio factor groups are decoupled from each other. 
The \mbox{EA-CoRL} method maximizes the whole-body design for the chin-up motion, effectively utilizing the lower body to generate momentum while minimizing the load on the shoulder actuators. 

\begin{figure}[t]
	\centering
	\includegraphics[height=8.5cm]{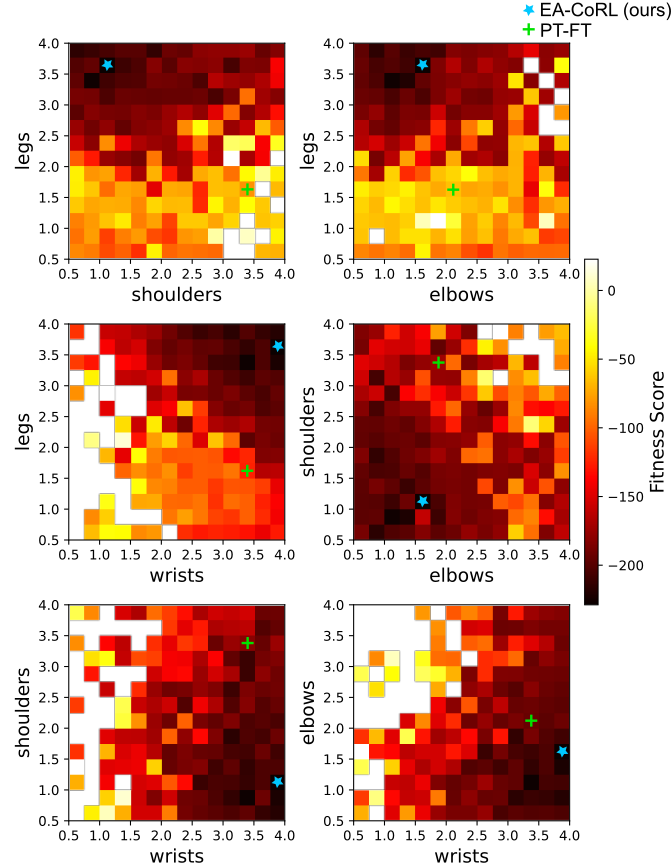}
	\caption{Heat-maps of RH5 gear-ratio factor's correlations for chin-up task. Blue stars represent the best design using the \mbox{EA-CoRL} method and green stars represent the best design using the PT-FT method. }
	\label{fig:heatmap2}
\end{figure}
Figure \ref{fig:heatmap2} visualizes the correlations between gear-ratio factor groups using six 2D heatmaps of fitness score distributions across the design space. 
Each cell represents the average fitness score over 50 \mbox{EA-CoRL} iterations in 4000 parallel simulation environments for a given design instance. 
The blue star indicates the gear ratio factors of the best design obtained using the \mbox{EA-CoRL} method. 
The green cross represents the best design obtained with the PT-FT method, allowing direct comparison of correlation differences between the four gear ratio factors. 
The heatmap is based on the fitness score distribution using \mbox{EA-CoRL}. 
The best PT-FT designs tend to maximize shoulder and wrist ratios but do not align with the high-reward regions (darker areas = higher reward). 
In contrast, the \mbox{EA-CoRL} designs fall within these high-reward regions, with more balanced shoulder and elbow ratios. 
Moreover, the best \mbox{EA-CoRL} design is consistently surrounded by high-reward regions, reinforcing its robustness. 

\begin{figure}[h]
	\centering
	\includegraphics[width =0.5\textwidth, height=5cm]{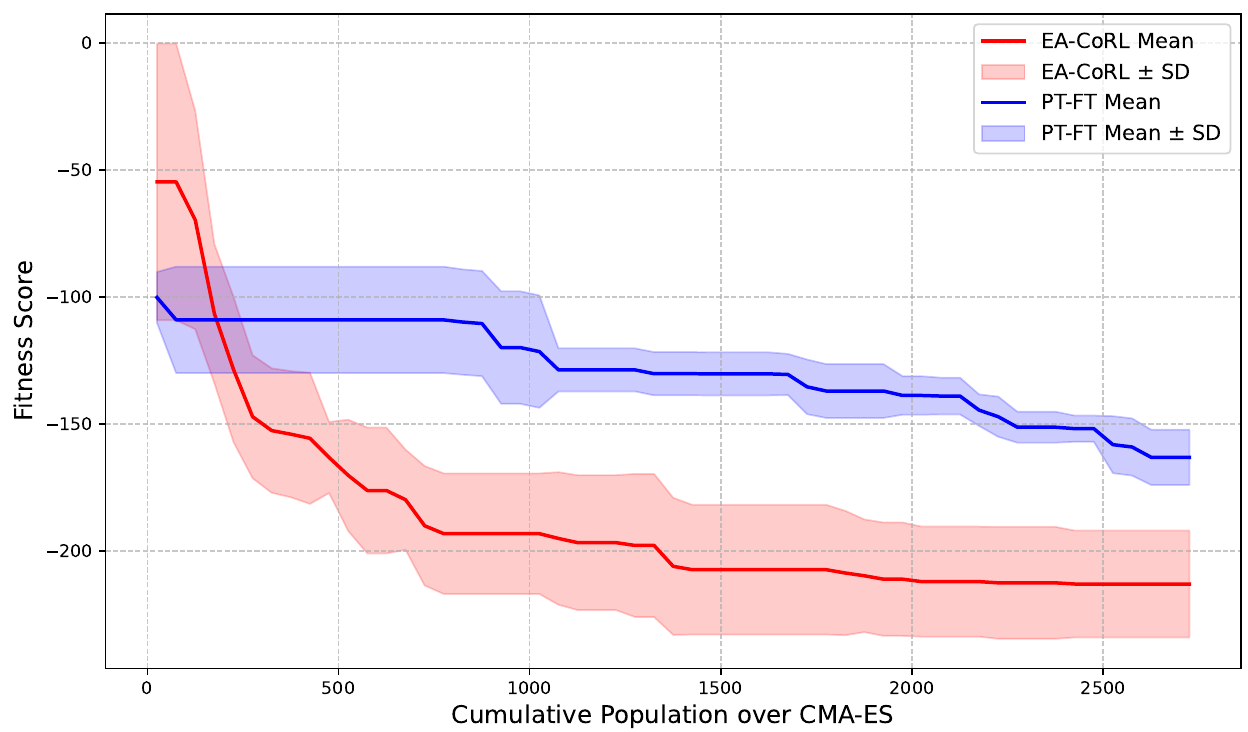}
	\caption{Comparison of \mbox{EA-CoRL} and PT-FT methods based on mean~$\pm$~standard deviation of fitness scores across the entire Design Evolution process, averaged over six random seeds. The x-axis represents the cumulative number of design's evolution.}
	\label{fig:fitness_function}
\end{figure}
Figure \ref{fig:fitness_function} shows the fitness score trajectories with standard deviations for \mbox{EA-CoRL} and PT-FT across the design evolution process, averaged over six seeds. 
The results indicate that \mbox{EA-CoRL} achieves a significantly lower final fitness score (higher reward), while PT-FT stagnates. 
This discrepancy is due to continuous policy updates in \mbox{EA-CoRL}, which allow more dynamic adaptation. 
These observations confirm that \mbox{EA-CoRL} enables a more effective balance between exploration and exploitation. 

\begin{figure}[h]
	\centering
	\includegraphics[width =0.45\textwidth, height=4cm]{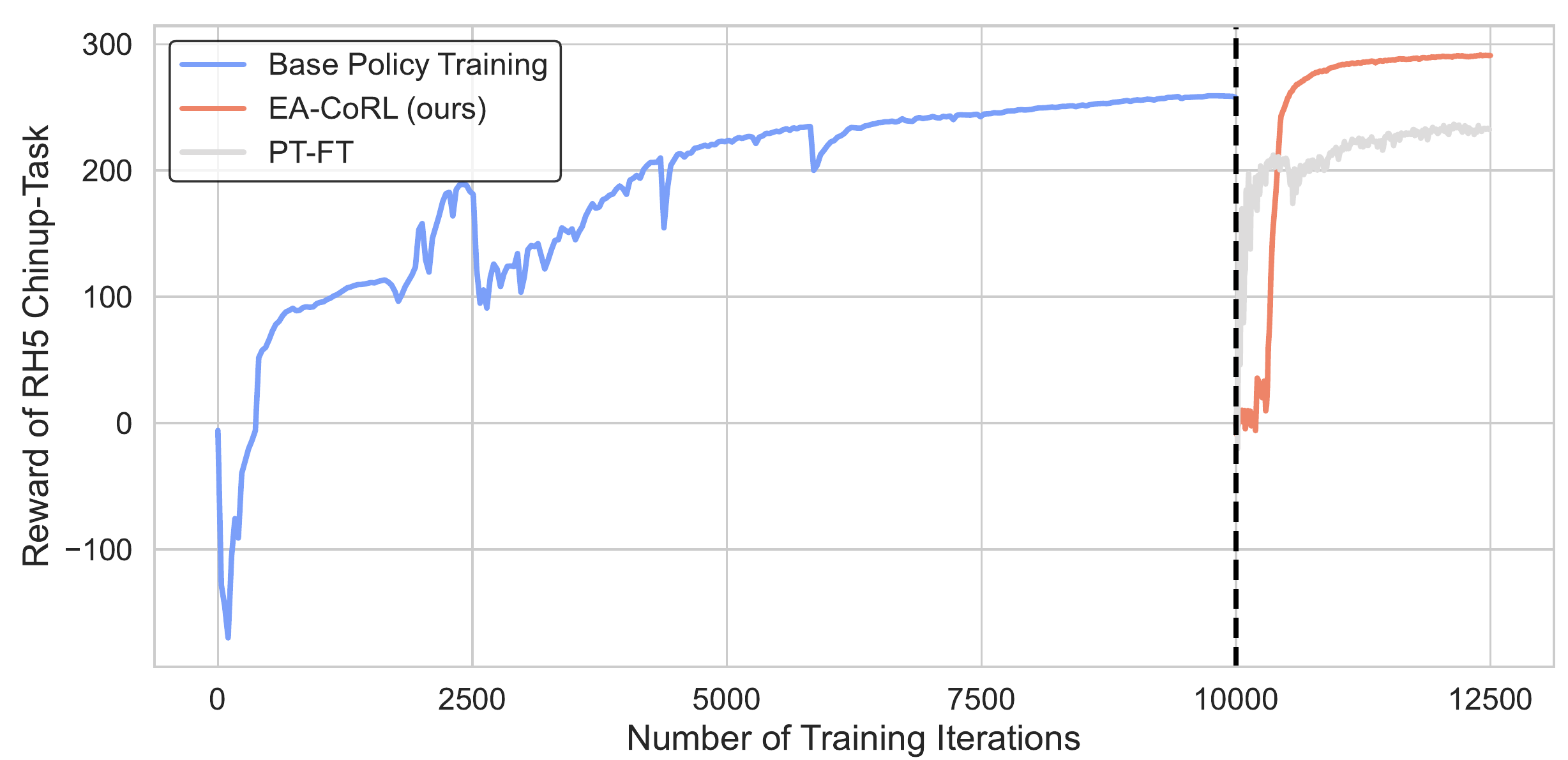}
	\caption{Learning curves for the RH5 chin-up task, comparing the base policy (blue), \mbox{EA-CoRL} (orange), and PT-FT (grey).}
	\label{fig:learning_curve}
\end{figure}
Figure \ref{fig:learning_curve} presents the RL curves for the \mbox{EA-CoRL} co-design applied to the RH5 chin-up task. 
The blue curve represents the training performance of the base policy \( \pi_0 \), the orange curve the best \mbox{EA-CoRL} design \( \pi_\theta \), and the gray curve the best PT-FT design. 
Training is extended to 10,000 iterations for fair comparison. 
\mbox{EA-CoRL} achieves the highest final task reward, benefiting from continuous adaptation. 
PT-FT, by contrast, converges quickly but remains frozen within the pre-trained policy region, preventing further improvement. 
Both methods, however, show rapid convergence compared to training independent policies from scratch, confirming the efficiency of fine-tuning from a pre-trained policy.

\subsection{Obtained best gear ratios}
Through the co-design process, \mbox{EA-CoRL} determined new gear ratio factors for both linear (screw pitches) and rotational (HarmonicDrive gear) actuators. When applied, these factors yield the effective joint gear ratios summarized in~\tref{tab:gear_ratios}, and compared to the original RH5 design and the PT-FT best design. Our new gear ratios reveal substantial adjustments that can be implemented without a meaningful increase in system mass. 
%
%
\begin{table}[h!]
\centering
\caption{\mbox{EA-CoRL} derived best gear ratios vs. original RH5 and the best obtained using PT-FT (smaller screw pitches increase torque but reduce speed).}
\label{tab:gear_ratios}
\begin{tabular}{lccc}
\\[-15pt]
\hline
\textbf{Actuator} & \textbf{\mbox{EA-CoRL}} & \textbf{Original} & \textbf{PT-FT} \\
\hline
Shoulder (rotational)& $1{:}112$  & $1{:}100$  & $1{:}349$ \\  
Elbow (linear) & $1.17\,\mathrm{mm}$  &$2.00\,\mathrm{mm}$ & $0.96\,\mathrm{mm}$ \\  
Wrist pitch/yaw (linear)& $0.26\,\mathrm{mm}$ & $1.00\,\mathrm{mm}$ & $0.29\,\mathrm{mm}$\\ 
Body pitch (linear)& $1.42\,\mathrm{mm}$ & $5.00\,\mathrm{mm}$ & $2.94\,\mathrm{mm}$ \\
Hip pitch (linear) & $2.85\,\mathrm{mm}$ & $10.00\,\mathrm{mm}$ & $5.88\,\mathrm{mm}$\\
Knee (linear)  & $2.85\,\mathrm{mm}$ & $10.00\,\mathrm{mm}$ & $5.88\,\mathrm{mm}$\\
\hline
\end{tabular}
\end{table}
\subsection{Discussion} 
The results demonstrate that the \mbox{EA-CoRL} method performs better than the PT-FT method, achieving higher final task rewards and exhibiting greater exploration capacity.
A key aspect of our analysis is the evaluation of the best gear ratios obtained through \mbox{EA-CoRL}. The results show that the shoulder actuators’ gear ratios remain close to the original design (1:100), with the best value being 1:112. Interestingly, the shoulder yaw actuator operates at only 18\% of its torque limit compared to the shoulder roll and pitch actuators, indicating its limited contribution to the chin-up motion. In contrast, the higher gear ratios in the wrist and leg actuators emphasize their critical roles in executing the task. This suggests that, unlike humans who rely on a flexible and powerful back muscle system, RH5 compensates for structural limitations by increasing torque demands in the wrists and lower body to successfully perform the chin-up motion.
\mbox{EA-CoRL}'s advantage stems from its continuous policy updates, which enable adaptive RL training. By continuously refining the control policy in response to design changes, \mbox{EA-CoRL} facilitates more flexible and efficient exploration of the design space. This approach enables superior online RL performance compared to PT-FT, which adapts policies only from a fixed pre-trained baseline PT-FT employs a bi-level strategy where an evolutionary algorithm refines design solutions while the control policy remains tied to the same pre-trained model. Although this structure allows PT-FT to achieve faster convergence in the early stages of co-design, it also risks premature stagnation. In contrast, \mbox{EA-CoRL}'s continuous adaptation mechanism promotes broader exploration.


\section{Conclusion and Outlooks}
\label{sec:conclusion} 
This work introduced \mbox{EA-CoRL}, applied to the whole-body humanoid RH5 for a highly dynamic chin-up task. By categorizing co-design variables into four groups, we reduced the dimensionality of the state space while influencing 17 DoFs across joint velocities and torque limits, yielding 34 design parameters per configuration. 
The proposed bi-level framework, which combines CMA-ES for design sampling with parallelized RL training in a dynamic simulator, demonstrated the advantage of continuous policy adaptation to varying design configurations. Results confirm EA-CoRL’s ability to efficiently explore both robot design and control policy, with strong potential for real-world deployment. Its model-agnostic nature further supports applicability to other humanoid and legged robot co-design problems.
The findings underscore the importance of integrating continuous policy adaptation directly into the co-design process. By jointly optimizing robot morphology and control, \mbox{EA-CoRL} ensures adaptability to evolving design constraints and produces more versatile solutions.
The current approach relies solely on reward as the design metric, yet encoding all relevant physical factors into a single reward function is inherently difficult and prone to reward engineering. As future work, we plan to extend this framework toward multi-task optimization, training a humanoid on a benchmark set of humanoid-specific tasks. 
We also aim to investigate reward formulations inspired by biomechanical theories, and to develop performance metrics that better capture the design space. 
Incorporating advanced optimization techniques such as Bayesian optimization, and learning paradigms such as imitation learning, could further improve benchmarking and refinement.
The main limitation lies in computational cost, driven primarily by convex hull collision computations between the high-bar and hollow cylinders in dynamic simulation, which also introduce training instabilities. More efficient algorithms for these computations would enhance scalability and real-world applicability, helping to bridge the sim-to-real gap.
Addressing these challenges will make \mbox{EA-CoRL} a more general and powerful co-design framework, advancing humanoid robotics for dynamic real-world applications.
\bibliographystyle{IEEEtran}
\begin{samepage}
\bibliography{references}
\end{samepage}

%
%

\end{document}